\documentclass{article}
\usepackage{ijcai24}
\usepackage{enumitem}
\usepackage{times}
\usepackage{soul}
\usepackage{url}
\usepackage[hidelinks]{hyperref}
\usepackage[utf8]{inputenc}
\usepackage[small]{caption}
\usepackage{graphicx}
\usepackage{amsmath}
\usepackage{amsthm}
\usepackage{booktabs}
\usepackage{multicol}
\usepackage{algorithm}
\usepackage{multirow}
\usepackage{algorithmic}
\usepackage[switch]{lineno}
\usepackage{xcolor}
\usepackage{natbib}
\definecolor{mygreen}{RGB}{0,150,0}
\definecolor{myred}{RGB}{255,0,0}

\usepackage[draft,textsize=footnotesize,textwidth=15mm]{todonotes}

\definecolor{paired-light-blue}{RGB}{198, 219, 239}
\definecolor{paired-dark-blue}{RGB}{49, 130, 188}
\definecolor{paired-light-orange}{RGB}{251, 208, 162}
\definecolor{paired-dark-orange}{RGB}{230, 85, 12}
\definecolor{paired-light-green}{RGB}{199, 233, 193}
\definecolor{paired-dark-green}{RGB}{49, 163, 83}
\definecolor{paired-light-purple}{RGB}{218, 218, 235}
\definecolor{paired-dark-purple}{RGB}{117, 107, 176}
\definecolor{paired-light-gray}{RGB}{217, 217, 217}
\definecolor{paired-dark-gray}{RGB}{99, 99, 99}
\definecolor{paired-light-pink}{RGB}{222, 158, 214}
\definecolor{paired-dark-pink}{RGB}{123, 65, 115}
\definecolor{paired-light-red}{RGB}{231, 150, 156}
\definecolor{paired-dark-red}{RGB}{131, 60, 56}
\definecolor{paired-light-yellow}{RGB}{231, 204, 149}
\definecolor{paired-dark-yellow}{RGB}{141, 109, 49}

\definecolor{bg1}{HTML}{FF9966}
\definecolor{bg2}{HTML}{CCE5FF}
\definecolor{bg3}{HTML}{FFCC99}
\definecolor{bg4}{HTML}{FFC107}
\definecolor{bg5}{HTML}{FFCCCC}
\definecolor{bg6}{HTML}{D5E8D4}
\definecolor{bg7}{HTML}{eeeeee}
\definecolor{bg8}{HTML}{cdeb8b}
\definecolor{bg9}{HTML}{dae8fc}
\definecolor{bg10}{HTML}{a2e6eb}

\definecolor{bg31}{HTML}{FFCDD2} 

\definecolor{bg32}{HTML}{F8BBD0}

\definecolor{bg33}{HTML}{E1BEE7} 

\definecolor{bg34}{HTML}{D7CCC8} 

\definecolor{bg35}{HTML}{B2DFDB} 

\definecolor{bg36}{HTML}{A5D6A7} 

\definecolor{bg37}{HTML}{FFF9C4} 

\definecolor{bg38}{HTML}{FFECB3} 

\definecolor{bg111}{HTML}{CB6843}

\definecolor{bg112}{HTML}{D77C5C}

\definecolor{bg113}{HTML}{E28E6E}
\definecolor{bg114}{HTML}{E89F7D}
\definecolor{bg115}{HTML}{EDAE8A}
\definecolor{bg116}{HTML}{F0BA95}
\definecolor{bg117}{HTML}{F3C29F}
\definecolor{bg118}{HTML}{F6CCAA}
\definecolor{bg119}{HTML}{F8D5B3}
\definecolor{bg120}{HTML}{FADCBD}
\definecolor{bg121}{HTML}{FCE6C7}

\definecolor{bg39}{HTML}{FFE0B2} 

\definecolor{bg40}{HTML}{3CB371} 

\definecolor{bg43}{HTML}{ffe5d9}

\definecolor{bg15}{HTML}{7FFFD4}

\definecolor{bg17}{HTML}{F0FFFF}

\definecolor{bg18}{HTML}{F5FFFA}

\definecolor{bg19}{HTML}{F8F8FF}

\definecolor{bg20}{HTML}{FFFFFF}

\definecolor{bg21}{HTML}{E1F5FE}

\definecolor{bg22}{HTML}{B3E5FC}

\definecolor{bg23}{HTML}{81D4FA}

\definecolor{bg24}{HTML}{4FC3F7}

\definecolor{bg25}{HTML}{29B6F6}

\definecolor{bg26}{HTML}{03A9F4}

\definecolor{bg27}{HTML}{039BE5}

\definecolor{bg28}{HTML}{0288D1}

\definecolor{bg29}{HTML}{0277BD}

\definecolor{bg30}{HTML}{01579B}

\definecolor{bg16}{HTML}{FFCC99}

\definecolor{pg51}{HTML}{E8F5E9} 
\definecolor{pg52}{HTML}{C8E6C9} 
\definecolor{pg53}{HTML}{B9F6CA} 
\definecolor{pg54}{HTML}{A9DFBF} 
\definecolor{pg55}{HTML}{BCF5A6} 

\definecolor{pg56}{HTML}{BEF1CE} 
\definecolor{pg57}{HTML}{CEF6EC} 
\definecolor{pg58}{HTML}{B7F0B1} 
\definecolor{pg59}{HTML}{B1F2B5} 
\definecolor{pg60}{HTML}{9DF3C4} 

\definecolor{pg61}{HTML}{DEF7E0} 
\definecolor{pg62}{HTML}{E8F8DC} 

\definecolor{pg63}{HTML}{EBF7E7} 
\definecolor{pg64}{HTML}{F0FDF4} 

\definecolor{pg65}{HTML}{F1FEE7} 
\definecolor{pg66}{HTML}{F7FFF6} 
\definecolor{pg67}{HTML}{FCFFE7} 
\definecolor{pg68}{HTML}{F4FFD2} 
\definecolor{pg69}{HTML}{EEFFE2} 
\definecolor{pg70}{HTML}{E3FDF5} 

\definecolor{connect-color}{RGB}{0,0,0}
\definecolor{middle-color}{RGB}{255,255,255}
\definecolor{leaf-color}{RGB}{173,216,230}
\definecolor{line-color}{RGB}{25,25,112}

\usepackage{url}            
\usepackage{booktabs}       
\usepackage{amsfonts}       
\usepackage{nicefrac}       
\usepackage{microtype}      
\usepackage{xcolor}         
\usepackage{amsmath}
\usepackage{graphicx}
\usepackage{enumitem}
\usepackage{multirow}
\usepackage{subcaption}
\usepackage{listings}
\usepackage{tcolorbox}
\usepackage{xspace}
\usepackage{makecell}
\usepackage{soul}               
\setstcolor{red}            
\usepackage{tikz}

\usepackage{todonotes}

\usepackage{longtable}
\usepackage{tablefootnote}
\usepackage[edges]{forest}
\definecolor{hidden-draw}{RGB}{20,68,106}
\definecolor{hidden-pink}{RGB}{255,245,247}
\definecolor{red}{RGB}{255,0,0}

\definecolor{hidden-draw}{RGB}{0,0,0}
\definecolor{hidden-pink}{RGB}{255,182,193}

\usepackage[T1]{fontenc}

\usepackage[utf8]{inputenc}

\usepackage{microtype}

\usepackage{inconsolata}

\usepackage{forest}    
\usepackage{hyperref}  
\usepackage{tikz}      

\tikzset{
    root style/.style={
        draw,
        rounded corners,
        fill=blue!30, 
        align=center,
        font=\bfseries
    },
    child style/.style={
        draw,
        rounded corners,
        fill=green!30, 
        align=center,
        font=\bfseries
    },
    grandchild style/.style={
        draw,
        rounded corners,
        fill=red!30, 
        align=center,
        font=\bfseries
    }
}

\tikzset{
  my-box/.style={
    rectangle,
    draw=hidden-draw,
    rounded corners,
    text opacity=1,
    minimum height=1.5em,
    minimum width=40em,
    inner sep=2pt,
    align=center,
    line width=0.8pt,
  },
  leaf/.style={
    my-box,
    minimum height=1.5em,
    text=black,
    align=center,
    font=\normalsize,
    inner xsep=2pt,
    inner ysep=4pt,
    line width=0.8pt,
  }
}

\title{A Systematic Survey of Prompt Engineering in Large Language Models: Techniques and Applications}

\author{
Pranab Sahoo$^1$
\and
Ayush Kumar Singh$^1$\and
Sriparna Saha$^1$\and
Vinija Jain$^{2,3}$\footnote{Work does not relate to position at Amazon.}\and
Samrat Mondal$^1$\And
Aman Chadha$^{2,3}$\textsuperscript{*}\\
\affiliations
$^1$Department of Computer Science And Engineering, Indian Institute of Technology
Patna\\
$^2$Stanford University, $^3$Amazon AI\\
\emails{
\texttt{\{pranab\_2021cs25, ayush\_2211ai27, sriparna, samrat\}@iitp.ac.in,
hi@vinija.ai,
hi@aman.ai}
} }

\begin{document}

\maketitle

\begin{abstract}
Prompt engineering has emerged as an indispensable technique for extending the capabilities of large language models (LLMs) and vision-language models (VLMs). This approach leverages task-specific instructions, known as prompts, to enhance model efficacy without modifying the core model parameters. Rather than updating the model parameters, prompts allow seamless integration of pre-trained models into downstream tasks by eliciting desired model behaviors solely based on the given prompt. Prompts can be natural language instructions that provide context to guide the model or learned vector representations that activate relevant knowledge. This burgeoning field has enabled success across various applications, from question-answering to commonsense reasoning. However, there remains a lack of systematic organization and understanding of the diverse prompt engineering methods and techniques. This survey paper addresses the gap by providing a structured overview of recent advancements in prompt engineering, categorized by application area. For each prompting approach, we provide a summary detailing the prompting methodology, its applications, the models involved, and the datasets utilized. We also delve into the strengths and limitations of each approach and include a taxonomy diagram and table summarizing datasets, models, and critical points of each prompting technique. This systematic analysis enables a better understanding of this rapidly developing field and facilitates future research by illuminating open challenges and opportunities for prompt engineering.

\end{abstract}

\begin{figure}
  \centering
     \includegraphics[width =0.50\textwidth, height=3cm]{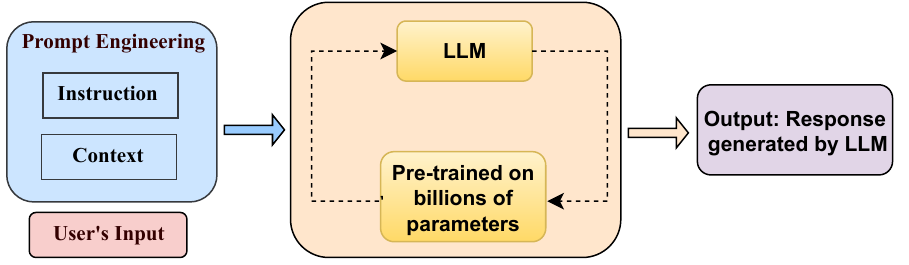}\hfill
     \caption{Visual breakdown of prompt engineering components: LLMs trained on extensive data, instruction and context as pivotal elements shaping the prompt, and a user input interface.}
  \label{fig:proposedfedbndp}
\end{figure}

\section{Introduction}
\begin{figure*}[ht!]
  \centering
  \resizebox{0.85\textwidth}{!}{
    \begin{forest}
      forked edges,
      for tree={
        grow=east,
        reversed=true,
        anchor=base west,
        parent anchor=east,
        child anchor=west,
        base=center,
        font=\large,
        rectangle,
        draw=hidden-draw,
        rounded corners,
        align=center,
        text centered,
        minimum width=5em,
        edge+={darkgray, line width=1pt},
        s sep=3pt,
        inner xsep=2pt,
        inner ysep=3pt,
        line width=0.8pt,
        ver/.style={rotate=90, child anchor=north, parent anchor=south, anchor=center},
      },
      where level=1{text width=15em,font=\normalsize,}{},
      where level=2{text width=14em,font=\normalsize,}{},
      where level=3{minimum width=10em,font=\normalsize,}{},
      where level=4{text width=26em,font=\normalsize,}{},
      where level=5{text width=20em,font=\normalsize,}{},
      [
        \textbf{Prompt Engineering}, for tree={fill=paired-light-red!80}, text width=14em
        [
              \textbf{New Tasks Without Extensive}\\ \textbf{Training} \S\ref{newtask}, for tree={fill=red!50}, text width=18em
              [
                \textbf{Zero-shot Prompting} \citep{radford2019language}, for tree={fill=pg58},text width=28em
              ]
              [
                \textbf{Few-shot Prompting} \citep{brown2020language}, for tree={fill=pg58},text width=28em
              ]
        ]
        [
            \textbf{Reasoning and Logic} \S\ref{ral}, for tree={fill=green!50}, text width=18em
            [
            \textbf{Chain-of-Thought (CoT) Prompting}~\citep{wei2022chain} , for tree={fill=bg4},text width=28em
            ]
            [
            \textbf{Automatic Chain-of-Thought (Auto-CoT)}~\citep{zhang2022automatic} , for tree={fill=bg4},text width=28em
            ]
            [
            \textbf{Self-Consistency}~\citep{wang2022self} , for tree={fill=bg4},text width=28em
            ]
            [
            \textbf{Logical CoT (LogiCoT) Prompting}~\citep{zhao2023enhancing} , for tree={fill=bg4},text width=28em
            ]p
            [
            \textbf{Chain-of-Symbol (CoS) Prompting}~\citep{hu2023chainofsymbol} , for tree={fill=bg4},text width=28em
            ]
            [
            \textbf{Tree-of-Thoughts (ToT) Prompting}~\citep{yao2023tree} , for tree={fill=bg4},text width=28em
            ]
            [
            \textbf{Graph-of-Thought (GoT) Prompting}~\citep{yao2023beyond} , for tree={fill=bg4},text width=28em
            ]
            [
            \textbf{System 2 Attention Prompting}~\citep{weston2023system} , for tree={fill=bg4},text width=28em
            ]
            [
            \textbf{Thread of Thought (ThoT) Prompting}~\citep{zhou2023thread} , for tree={fill=bg4},text width=28em
            ]
            [
            \textbf{Chain of Table Prompting}~\citep{wang2024chainoftable} , for tree={fill=bg4},text width=28em
            ]
            [
            \textbf{ Self-Refine Prompting}~\citep{madaan2023selfrefine} , for tree={fill=bg4},text width=28em
            ] 
            [
            \textbf{ Code Prompting}~\citep{puerto2024codepromptingelicitsconditional} , for tree={fill=bg4},text width=28em
            ]
            [
            \textbf{Self-Harmonized CoT (ECHO) Prompting}~\citep{mekala2024echopromptinstructingmodelrephrase} , for tree={fill=bg4},text width=28em
            ]
            [
            \textbf{Logic-of-Thought Prompting}~\citep{liu2024logicofthoughtinjectinglogiccontexts} , for tree={fill=bg4},text width=28em
            ]
            [
            \textbf{ Instance-adaptive Prompting (IAP) }~\citep{yuan2024instanceadaptivezeroshotchainofthoughtprompting} , for tree={fill=bg4},text width=28em
            ]
            [
            \textbf{ End-to End DAG-Path (EEDP) Prompting }~\citep{yuan2024instanceadaptivezeroshotchainofthoughtprompting} , for tree={fill=bg4},text width=28em
            ]
            [
            \textbf{ Layer-of-Thoughts (LoT) }~\citep{fungwacharakorn2024layerofthoughtspromptinglotleveraging} , for tree={fill=bg4},text width=28em
            ]
            [
            \textbf{ Narrative-of-Thought (NoT) Prompting}~\citep{zhang2024narrativeofthoughtimprovingtemporalreasoning} , for tree={fill=bg4},text width=28em
            ]
            [
             \textbf{ Buffer of Thoughts (BoT) Prompting}~\citep{yang2024buffer} , for tree={fill=bg4},text width=28em
            ]
            [
            \textbf{ Contrastive Denoising with Noisy Chain-of-Thought (CD-CoT)}\\ \textbf{Prompting}~\citep{NEURIPS2024_dfaa29ed} , for tree={fill=bg4},text width=28em
            ]
            [
            \textbf{Reverse Chain-of-Thought (R-CoT) Prompting}~\citep{deng2024rcotreversechainofthoughtproblem} , for tree={fill=bg4},text width=28em
            ]
            [
            \textbf{Chain of Draft (CoD) Prompting}~\citep{xu2025chaindraftthinkingfaster} , for tree={fill=bg4},text width=28em
            ]   
        ]
        [
            \textbf{Reduce Hallucination} \S\ref{rh}, for tree={fill=lime!50}, text width=18em
            [
            \textbf{Retrieval Augmented Generation (RAG)}~\citep{lewis2020retrieval} , for tree={fill=bg32},text width=28em
            ]
            [
            \textbf{ReAct Prompting}~\citep{yao2022react} , for tree={fill=bg32},text width=28em
            ]
            [
            \textbf{Chain-of-Verification (CoVe)}~\citep{dhuliawala2023chain} , for tree={fill=bg32},text width=28em
            ]
            [
            \textbf{Chain-of-Note (CoN) Prompting}~\citep{yu2023chainofnote} , for tree={fill=bg32},text width=28em
            ]
            [
            \textbf{Chain-of-Knowledge (CoK) Prompting}~\citep{li2023chainofknowledge} , for tree={fill=bg32},text width=28em
            ]
        ]
        [
            \textbf{User Interaction} \S\ref{ui}, for tree={fill=bg25}, text width=18em
            [
            \textbf{Active-Prompt}~\citep{diao2023active} , for tree={fill=bg8},text width=28em
            ]
        ]
        [
            \textbf{Fine-Tuning and Optimization} \S\ref{fto}, for tree={fill=purple!50}, text width=18em
            [
            \textbf{Automatic Prompt Engineer (APE)}~\citep{zhou2022large} , for tree={fill=bg10},text width=28em
            ]
        ]
        [
            \textbf{Knowledge-Based Reasoning and} \\ \textbf{Generation}\S\ref{kbrg}, for tree={fill=bg34}, text width=18em
            [
            \textbf{Automatic Reasoning}\\ \textbf{and Tool-use (ART)}~\citep{paranjape2023art} , for tree={fill=bg37},text width=28em
            ]
        ]
        [
            \textbf{Improving Consistency} \\ \textbf{and Coherence} \S\ref{icc}, for tree={fill=bg33}, text width=18em
            [
            \textbf{Contrastive Chain-of-Thought} \\ \textbf{Prompting (CCoT)}~\citep{chia2023contrastive} , for tree={fill=pink!50},text width=28em
            ]
        ]
        [
            \textbf{Managing Emotions and Tone} \S\ref{met}, for tree={fill=bg40}, text width=18em
            [
            \textbf{Emotion Prompting}~\citep{li2023large} , for tree={fill=paired-light-yellow!50},text width=28em
            ]
        ]
        [
          \textbf{Code Generation and Execution} \S\ref{cge}, for tree={fill=cyan!50}, text width=18em
              [
                \textbf{Scratchpad Prompting}~\citep{nye2021show} , for tree={fill=bg16},text width=28em
              ]
              [
                \textbf{Program of Thoughts (PoT) Prompting}~\citep{chen2022program},text width=28em, for tree={fill=bg16}
                ]
              [
                \textbf{Structured Chain-of-Thought}\\ \textbf{(SCoT) Prompting}~  \citep{li2023structured},text width=28em, for tree={fill=bg16}
                ]
             [
                \textbf{Chain of Code (CoC) Prompting}~\citep{li2023chain},text width=28em, for tree={fill=bg16}
                ]
        ]
        [
            \textbf{Optimization and Efficiency} \S\ref{oe}, for tree={fill=bg15}, text width=18em
            [
                \textbf{Optimization by Prompting}~\citep{yang2023large}, text width=28em, for tree={fill=bg7}
            ]
        ]
        [
            \textbf{Understanding User Intent} \S\ref{uui}, for tree={fill=blue!50}, text width=18em
            [
                \textbf{Rephrase and Respond (RaR) Prompting}~\citep{deng2023rephrase}, text width=28em, for tree={fill=bg38} 
            ]
        ]
        [
            \textbf{Metacognition and Self-Reflection} \S\ref{mas}, for tree={fill=bg1}, text width=18em
            [
                \textbf{Take a Step Back Prompting}~\citep{zheng2023take}, text width=28em, for tree={fill=bg36}
            ]
        ]
      ]
    \end{forest}
    }
  \caption{Taxonomy of prompt engineering techniques in LLMs, organized around application domains, providing a nuanced framework for customizing prompts across diverse contexts.}
  \label{fig:lit_surv}
\end{figure*}
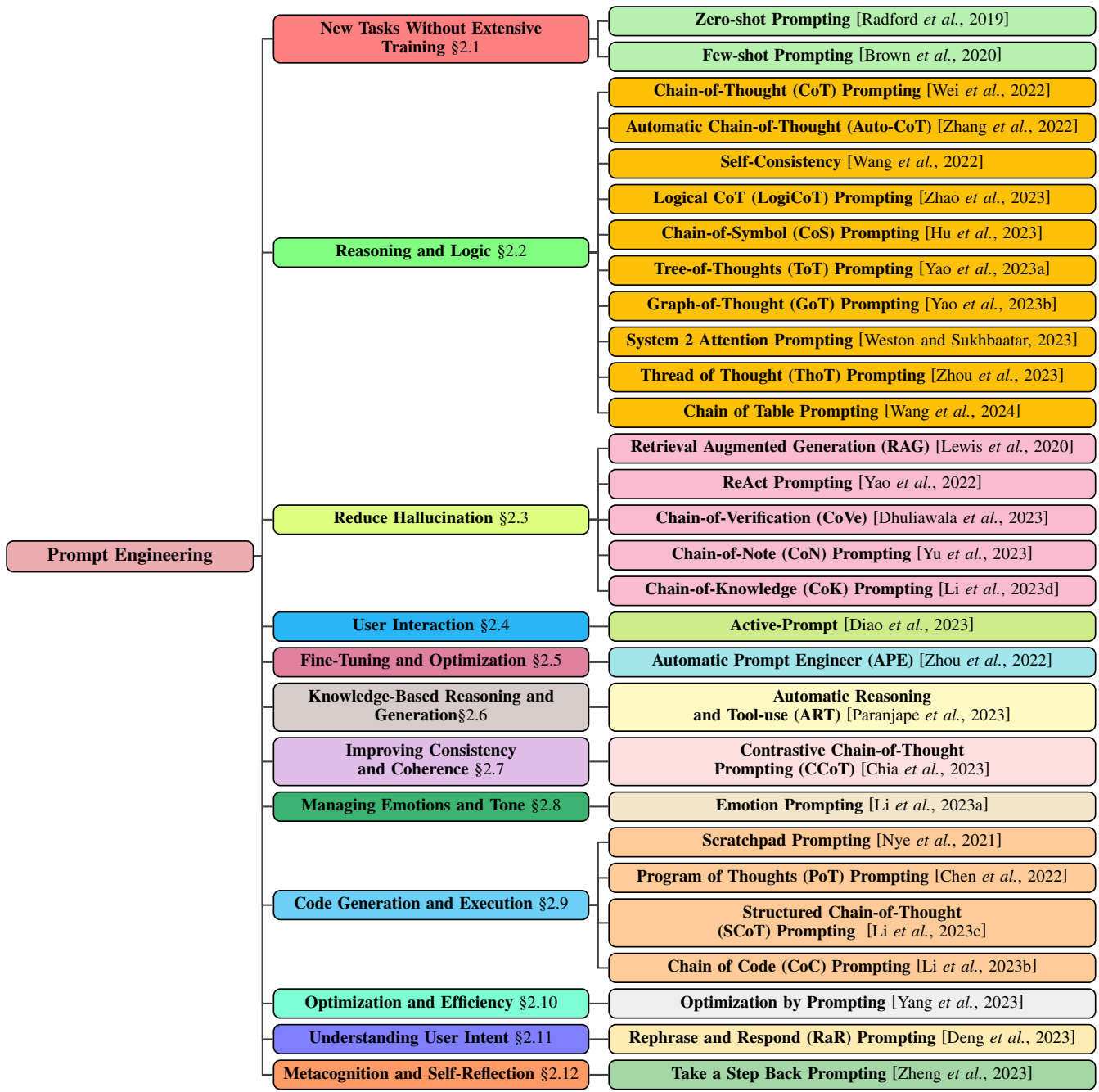

Prompt engineering has emerged as a crucial technique for enhancing the capabilities of pre-trained large language models (LLMs) and vision-language models (VLMs). It involves strategically designing task-specific instructions, referred to as prompts, to guide model output without altering parameters. The significance of prompt engineering is especially evident in its transformative impact on the adaptability of LLMs and VLMs. By offering a mechanism to fine-tune model outputs through carefully crafted instructions, prompt engineering enables these models to excel across diverse tasks and domains. This adaptability is different from traditional paradigms, where model retraining or extensive fine-tuning is often required for task-specific performance. This is the transformative promise of prompt engineering, pushing the boundaries of AI and opening doors to a future brimming with possibilities. In an ever-evolving landscape, ongoing research consistently reveals innovative approaches and applications within prompt engineering. The significance of prompt engineering is underscored by its capacity to steer model responses, enhancing the adaptability and applicability of LLMs across diverse sectors. The landscape of contemporary prompt engineering spans a spectrum of techniques, encompassing foundational methods like zero-shot and few-shot prompting to more intricate approaches such as "chain of code" prompting. The notion of prompt engineering was initially investigated and popularized in the LLMs~\citep{liu2023pre},~\citep{tonmoy2024comprehensive},~\citep{chen2023unleashing} later extended to VLMs~\citep{wu2023visual},~\citep{bahng2022exploring}. 
Despite the extensive literature on prompt engineering within both LLMs and VLMs, a notable gap remains, particularly concerning a systematic overview of application-centric prompt engineering techniques. With recent strides in prompt engineering, there is a pressing need for a comprehensive survey that offers a nuanced understanding of applications and advancements in contemporary research. This survey dives deep into the ever-evolving landscape of prompt engineering, analyzing over 41 distinct techniques categorized by their diverse applications. Employing a systematic review approach, we meticulously delve into the intricacies of diverse cutting-edge prompting methods. Our examination encompasses their applications, the language models utilized, and the datasets subjected to experimentation, providing a detailed and nuanced analysis of the evolving landscape of prompt engineering. Additionally, we discuss the pros and cons of these techniques, offering insights into their comparative efficacy.
We present a comprehensive taxonomy diagram that illustrates how these techniques navigate the vast landscape of LLM capabilities (see Fig.\ref{fig:lit_surv}) and provide a table summarizing the datasets, employed models, and evaluation metrics (see Table\ref{table}). From language generation and question answering to code creation and reasoning tasks, prompt engineering empowers the LLMs into performing feats we never thought possible. By bridging the existing gap in the literature, this survey aims to serve as a valuable resource for researchers and practitioners, offering insights into the latest developments and facilitating a deeper understanding of the evolving landscape of prompt engineering. The structure of the paper is organized as follows: Section 2 presents the prompt engineering techniques from both basic to advanced by categorizing application-area and Section 3 provides a conclusion along with considerations for future research endeavors.

\section{Prompt Engineering}
In this section, we have organized prompt engineering techniques according to their application areas and provided a concise overview of the evolution of prompting techniques, spanning from zero-shot prompting to the latest advancements. 

\subsection{New Tasks Without Extensive Training}
\label{newtask}
\subsubsection{Zero-Shot Prompting} 
\label{zero}
Zero-shot prompting offers a paradigm shift in leveraging large LLMs. This technique removes the need for extensive training data, instead relying on carefully crafted prompts that guide the model toward novel tasks~\citep{radford2019language}. Specifically, the model receives a task description in the prompt but lacks labeled data for training on specific input-output mappings. The model then leverages its pre-existing knowledge to generate predictions based on the given prompt for the new task.



\subsubsection{Few-Shot Prompting}
\label{few}
Few-shot prompting provides models with a few input-output examples to induce an understanding of a given task, unlike zero-shot prompting, where no examples are supplied~\citep{brown2020language}. Providing even a few high-quality examples has improved model performance on complex tasks compared to no demonstration. However, few-shot prompting requires additional tokens to include the examples, which may become prohibitive for longer text inputs. Moreover, the selection and composition of prompt examples can significantly influence model behavior, and biases like favoring frequent words may still affect few-shot results. While few-shot prompting enhances capabilities for complex tasks, especially among large pre-trained models like GPT-3, careful prompt engineering is critical to achieve optimal performance and mitigate unintended model biases. 



\subsection{Reasoning and Logic}
\label{ral}
\subsubsection{Chain-of-Thought (CoT) Prompting} 
\label{cot}
LLMs often stumble in the face of complex reasoning, limiting their potential. Aiming to bridge this gap, \cite{wei2022chain} introduced Chain-of-Thought (CoT) prompting as a technique to prompt LLMs in a way that facilitates coherent and step-by-step reasoning processes. The primary contribution lies in the proposal and exploration of CoT prompting, demonstrating its effectiveness in eliciting more structured and thoughtful responses from LLMs compared to traditional prompts. Through a series of experiments, the authors showcase the distinctive qualities of CoT prompting, emphasizing its ability to guide LLMs through a logical reasoning chain. This results in responses that reflect a deeper understanding of the given prompts. For example, the prompt would show the reasoning process and final answer for a multi-step math word problem and mimic how humans break down problems into logical intermediate steps. The authors achieved state-of-the-art performance in math and commonsense reasoning benchmarks by utilizing CoT prompts for PaLM 540B, achieving an accuracy of 90.2\%.


\subsubsection{Automatic Chain-of-Thought (Auto-CoT) Prompting}
\label{auto-cot}
Manual creation of high-quality CoT examples is both time-consuming and suboptimal.~\cite{zhang2022automatic} introduced Auto-CoT to automatically instruct LLMs with a "Let's think step-by-step" prompt to generate reasoning chains. Recognizing the possibility of errors in individually generated chains, Auto-CoT enhances robustness through diverse sampling. It samples various questions and generates multiple distinct reasoning chains for each, forming a final set of demonstrations. This automated diverse sampling minimizes errors and enhances few-shot learning, eliminating the need for labor-intensive manual creation of reasoning chains. Auto-CoT demonstrated enhanced performance, surpassing the CoT paradigm with average accuracy improvements of 1.33\% and 1.5\% on arithmetic and symbolic reasoning tasks, respectively, employing GPT-3.


\subsubsection{Self-Consistency}
\label{selfcon}
\cite{wang2022self} introduced self-consistency, a decoding strategy enhancing reasoning performance compared to greedy decoding in CoT prompting. For complex reasoning tasks with multiple valid paths, self-consistency generates diverse reasoning chains by sampling from the language model's decoder. It then identifies the most consistent final answer by marginalizing these sampled chains. This approach capitalizes on the observation that problems requiring thoughtful analysis often entail greater reasoning diversity, leading to a solution. The combination of self-consistency and chain-of-thought prompting results in significant accuracy improvements across various benchmarks, such as 17.9\% on GSM8K, 11.0\% on SVAMP, 12.2\% on AQuA, 6.4\% on StrategyQA, and 3.9\% on ARC-challenge compared to the baseline chain-of-thought prompting.

\subsubsection{Logical Chain-of-Thought (LogiCoT) Prompting}

\textls[-10]{The ability to perform logical reasoning is critical for LLMs to solve complex, multi-step problems across diverse domains. Existing methods, like CoT prompting, encourage step-by-step reasoning but lack effective verification mechanisms.  \cite{zhao2023enhancing} proposes a Logical Chain-of-Thought (LogiCoT) prompting, a neurosymbolic framework that leverages principles from symbolic logic to enhance reasoning in a coherent and structured manner. Specifically, LogiCoT applies the concept of reductio ad absurdum to verify each step of reasoning generated by the model and provide targeted feedback to revise incorrect steps. LogiCoT can reduce logical errors and hallucinations through a think-verify-revise loop. 
Experimenting with Vicuna-33b and GPT-4, the findings underscore LogiCoT's notable enhancement of reasoning abilities, exhibiting improvements of 0.16\% and 1.42\% on the GSM8K dataset and 3.15\% and 2.75\% on the AQuA dataset compared to CoT, respectively.}






\subsubsection{Chain-of-Symbol (CoS) Prompting}
\label{cos}
\textls[0]{LLMs often struggle with tasks involving complex spatial relationships due to their reliance on natural language, which is susceptible to ambiguity and biases. To overcome this limitation, \cite{hu2023chainofsymbol} introduced CoS, employing condensed symbols instead of natural language. CoS provides distinct advantages: clear and concise prompts, heightened spatial reasoning for LLMs, and improved human interpretability. CoS suffers from challenges such as scalability, generalizability, integration with other techniques, and interpretability of LLM reasoning based on symbols. Notably, the implementation of CoS significantly elevates ChatGPT's performance, boosting accuracy from 31.8\% to an impressive 92.6\% on Brick World tasks. Moreover, CoS achieves up to a 65.8\% reduction in prompt tokens, streamlining the process while maintaining high accuracy.}

\subsubsection{Tree-of-Thoughts (ToT) Prompting}
\label{tot}
\textls[-10]{\cite{yao2023tree} and ~\cite{long2023large} proposed the Tree-of-Thoughts (ToT) framework to enhance prompting capabilities for complex tasks requiring exploration and look-ahead reasoning. ToT extends CoT prompting by managing a tree structure of intermediate reasoning steps, known as "thoughts". Each thought represents a coherent language sequence moving toward the final solution. This structure allows language models to deliberately reason by assessing the progress generated by thoughts in solving the problem. ToT integrates the model's abilities to produce and evaluate thoughts with search algorithms like breadth-first or depth-first search. This enables systematic exploration among reasoning chains, with a look-ahead to expand promising directions and to backtrack when solutions are incorrect. ToT excelled in the Game of 24 tasks, achieving a 74\% success rate compared to CoT's 4\%. Additionally, in word-level tasks, ToT outperformed CoT with a 60\% success rate versus 16\%.}








\subsubsection{Graph-of-Thoughts (GoT) Prompting}
\label{got}
\textls[-10]{The inherent non-linear nature of human thought processes challenges the conventional sequential approach of CoT prompting.~\cite{yao2023beyond} introduced the "Graph of Thoughts" prompting, a graph-based framework advancing traditional sequential methods to better align with the non-linear characteristics of human thinking. This framework permits dynamic interplay, backtracking, and evaluation of ideas, allowing the aggregation and combination of thoughts from various branches, departing from the linear structure of the tree of thoughts. The key contributions encompass modeling the reasoning process as a directed graph, offering a modular architecture with diverse transformation operations. The framework is presented as a versatile and dynamic approach to language model prompting, capturing the intricacies of human thought processes and enhancing model capabilities. The GoT reasoning model demonstrates substantial gains over the CoT baseline, improving accuracy by 3.41\% with T5-base and 5.08\% with T5-large on GSM8K. It also boosts accuracy over the state-of-the-art Multimodal-CoT by 6.63\% using T5-base and 1.09\% with T5-large on ScienceQA.}

\subsubsection{System 2 Attention (S2A) Prompting}
\label{s2a}
\textls[-10]{The soft attention mechanism in Transformer-based LLMs is prone to incorporating irrelevant context information, impacting token generation adversely. To address this,~\cite{weston2023system} proposed System 2 Attention (S2A), utilizing the reasoning abilities of LLMs to selectively attend to relevant portions by regenerating the input context. S2A employs a two-step process to enhance attention and response quality by employing context regeneration and response generation with refined context. The effectiveness of S2A is evaluated across various tasks, including factual QA, long-form generation, and math word problems. In factual QA, S2A attains an accuracy of 80.3\%, demonstrating a substantial enhancement in factuality. In long-form generation, it improves objectivity and receives a score of 3.82 out of 5.}

\subsubsection{Thread of Thought (ThoT) Prompting}
\label{thot}
\textls[-10]{\cite{zhou2023thread} presented Thread of Thought (ThoT), a prompting technique designed to enhance the reasoning abilities of LLMs within chaotic contexts. ThoT, inspired by human cognition, systematically examines extensive contexts into manageable segments for incremental analysis, employing a two-phase approach where the LLM first summarizes and examines each segment before refining the information for a final response. ThoT's flexibility shines as a versatile "plug-and-play" module, enhancing reasoning across different models and prompting methods. Evaluations on question answering and conversation datasets reveal substantial performance improvements of 47.20\% and 17.8\%, respectively, especially in chaotic contexts.}


\subsubsection{Chain-of-Table Prompting}
\label{cotp}
\textls[-10]{Approaches like CoT, PoT, and ToT represent reasoning steps through free-form text or code, which face challenges when dealing with intricate table scenarios. The study by~\cite{wang2024chainoftable} introduced a pioneering prompting technique named Chain-of-Table. This method uses step-by-step tabular reasoning by dynamically generating and executing common SQL/DataFrame operations on tables. The iterative nature of this process enhances intermediate results, empowering LLMs to make predictions through logically visualized reasoning chains. Significantly, Chain-of-Table consistently improves the performance of two benchmark tabular datasets by 8.69\% on TabFact and 6.72\% on WikiTQ, respectively.}
\subsubsection{Self-Refine Prompting}\label{Self-refine}
\noindent \textls[-10]{Self-Refine prompting, proposed by ~\cite{madaan2023selfrefine}, enhances LLM performance by iteratively refining outputs through self-generated feedback, mimicking human revision. While LLMs can handle a wide range of tasks, they often struggle with complex objectives, ambiguous goals, or multi-step reasoning, leading to initial responses with inaccuracies or flawed logic. Inspired by human iterative refinement, Self-Refine enables LLMs to improve their outputs through a structured three-step process: generating an initial response, prompting the model to critique its own output, and refining the response based on this feedback. This cycle continues until predefined stopping criteria are met, allowing the model to produce more accurate and contextually relevant results. Unlike traditional prompting methods, which rely solely on a single-step response, Self-Refine fosters incremental improvement, making it particularly effective for tasks requiring nuanced reasoning. Experimental results demonstrate significant performance gains, with GPT-4 improving by 8.7 points in code optimization, 13.9 points in code readability, and 21.6 points in sentiment reversal tasks, showcasing its potential to enhance the reasoning and adaptability of LLMs across various domains.}
\subsubsection{Code Prompting}\label{code}
\noindent \textls[-10]{Pre-training on code enhances the reasoning capabilities of LLMs, yet the underlying mechanisms driving this improvement remain poorly understood. To investigate this,~\cite{puerto2024codepromptingelicitsconditional} examines the impact of input representation on LLM reasoning, specifically exploring whether reformulating natural language (NL) problems into code can trigger conditional reasoning abilities. 
This led to the introduction of Code Prompting, a technique that reformulates NL tasks into structured code, enabling direct prompting of text+code LLMs without relying on external code execution. Experiments on three reasoning benchmarks, ConditionalQA, BoardgameQA, and ShARC, demonstrate that code prompts significantly outperform traditional text-based prompts. On average, GPT 3.5 achieved a performance gain of 8.42 F1 score, while Mistral showed an average improvement of 4.22 across the three datasets.}
\subsubsection{Self-Harmonized Chain-of-Thought (ECHO) Prompting}
 \label{ECHO}
\noindent While Chain-of-Thought prompting enhances reasoning in LLMs, methods like Auto-CoT, which automate demonstration generation, face challenges from misleading similarity (incorrect rationales in similar examples) and ineffective diversity (irrelevant or overly varied patterns). To address these issues,~\cite{mekala2024echopromptinstructingmodelrephrase} introduced ECHO, a self-harmonized prompting framework that unifies diverse reasoning paths into a coherent pattern, balancing automation with robustness. ECHO operates through three key stages: (1) Question Clustering, where Sentence-BERT embeddings and k-means group questions into clusters; (2) Demonstration Sampling, which selects representative questions from each cluster and generates rationales using Zero-Shot-CoT; and (3) Demonstration Unification, where rationales are iteratively refined using a dynamic prompting mechanism to align reasoning patterns. This process minimizes diversity-induced noise while retaining adaptability. ECHO surpassed Auto-CoT by an average of 2.8\% across 10 reasoning benchmarks (arithmetic, commonsense, symbolic) while demonstrating greater efficiency. It retained performance with 50\% fewer examples, showing only a -0.8\% dip compared to Few-Shot-CoT’s -1.3\% decline. The method also achieved 2.3\% gains over Auto-CoT in Mixtral-8x7B, though it remained behind GPT-3.5, a gap attributed to differences in the quality of reasoning rationales.
\subsubsection{Logic-of-thought Prompting}
\label{LoT}
\noindent \textls[-10]{LLMs often exhibit unfaithful reasoning, where the generated conclusions diverge from the intermediate reasoning steps. Logic-of-Thought prompting~\citep{liu2024logicofthoughtinjectinglogiccontexts} is a neuro‐symbolic framework developed to mitigate this issue by enriching prompts with logical information derived from propositional logic. LoT operates in three phases: (1) Logic Extraction, during which LLMs identify propositions and logical relationships from input texts; (2) Logic Extension, in which a Python-based module applies formal logical laws (e.g., contraposition) to infer additional expressions; and (3) Logic Translation, where the extended logic is rendered back into natural language and appended to the original prompt to ensure contextual fidelity. Moreover, Logic-of-thought is designed to integrate seamlessly with other prompting strategies such as CoT, Self-Consistency, and ToT prompting. Reported evaluations indicate that Logic-of-thought can improve CoT accuracy on the ReClor benchmark by 4.35\%, enhance CoT prompting with Self-Consistency on LogiQA by 5\%, and further boost ToT prompting performance on the ProofWriter dataset by 8\%. Additionally, by preserving natural language representations throughout the process, Logic-of-Thought avoids the symbolic extraction errors that can impair other neuro‐symbolic systems, such as SatLM.}
\subsubsection{Instance-adaptive Prompting (IAP)}
\label{IAP}
\noindent \citet{yuan2024instanceadaptivezeroshotchainofthoughtprompting} tackle the generalization constraints of static task-level prompts (e.g., "Let’s think step by step") in zero-shot CoT reasoning by introducing Instance-Adaptive Prompting (IAP), a saliency-driven framework designed to dynamically tailor prompts to individual instances. Through information flow analysis of attention layers, the authors identified distinct patterns: effective reasoning correlates with strong semantic flow from questions to prompts in shallow layers and from integrated question-prompt representations to rationales in deeper layers. In contrast, fragmented or weak flows are indicative of suboptimal reasoning performance. IAP optimizes reasoning fidelity through two adaptive strategies. The first, IAP-ss (Sequential Substitution), enhances efficiency by iteratively testing prompts until predefined saliency thresholds are met. The second, IAP-mv (Majority Vote), prioritizes robustness by aggregating saliency scores across multiple prompts to determine consensus answers. Empirical evaluations underscore the broad applicability of IAP: in mathematical reasoning tasks (GSM8K, SVAMP), IAP-mv boosts the performance of LLaMA-3-8B and Qwen-14B by +1.82\% and +3.31\%, respectively, compared to static prompts. It achieves 19.25\% accuracy on causal judgment tasks, outperforming baselines at 16.04\%, and surpasses Self-Discover by +21.7\% on MMLU commonsense reasoning with Qwen-14B.
\subsubsection{End-to End DAG-Path (EEDP) Prompting }
\label{EEDP}
\noindent End-to-End DAG-Path (EEDP) prompting~\citep{hong2024endtoendgraphflatteningmethod} addresses the limitations of traditional graph-flattening methods, such as adjacency lists and edge lists, which struggle with long-distance reasoning in graph-related tasks for LLMs. EEDP's key insight is that conventional flattened representations often lose critical long-range dependencies essential for effective reasoning. To mitigate this, EEDP prioritizes the main backbone paths connecting graph endpoints (nodes with zero in-degree or out-degree) while preserving adjacency lists to maintain local contextual information. The EEDP framework operates through three key stages: (1) preprocessing input graphs into directed acyclic graphs (DAGs) using breadth-first search (BFS) to eliminate cycles, (2) extracting hierarchical paths between endpoints, and (3) compressing shared path segments with a differential pointer algorithm, effectively reducing token length by 55\% on molecular graphs. EEDP was evaluated on tasks such as Edge Prediction Connectivity Prediction (EPCP) and Edge Prediction Distance Prediction (EPDP) using educational (\texttt{Merged\_1000}) and molecular (\texttt{ZINC\_test\_2500}) datasets. The evaluation results highlighted significant performance gains over traditional baselines, with EPCP showing a +10.21\% accuracy improvement on \texttt{Merged\_1000} and +16.76\% on \texttt{ZINC\_test\_2500}. Similarly, EPDP achieved a +4.73\% accuracy boost on \texttt{Merged\_1000} and an impressive +30.13\% on \texttt{ZINC\_test\_2500}.
\subsubsection{Layer-of-Thoughts (LoT) Prompting}
\label{lot}
\noindent LLMs demonstrate strong performance in many reasoning tasks yet frequently face challenges with the precision–recall trade-off and explainability, particularly in complex legal retrieval scenarios. Layer-of-Thoughts (LoT) prompting~\citep{fungwacharakorn2024layerofthoughtspromptinglotleveraging} introduces a hierarchical framework that leverages constraint hierarchies to structure the reasoning process, thereby enhancing both retrieval accuracy and interpretability. In the context of legal document retrieval, LoT organizes reasoning into "layer thoughts" (conceptual stages) and "option thoughts" (partial solutions), applying sequential constraints to iteratively filter and refine candidate responses. For instance, the framework employs a three-layer process: (1) a Keyword Filtering Layer (KFL) that extracts LLM-generated keywords to initially filter documents using metrics such as at-least-k; (2) a Semantic Filtering Layer (SFL) that prioritizes documents based on multi-level relevance criteria and aggregation metrics; and (3) a Final Confirmation Layer (FCL) that validates the remaining candidates against the original query. By integrating both hard constraints (required) and soft constraints (preferential), LoT not only delivers explainable reasoning but also outperforms state-of-the-art models, for example, achieving an F2 score of 0.835 (with precision of 0.838 and recall of 0.839) on Japanese Civil Law retrieval compared to 0.807 for JNLP, and reaching near-perfect recall (0.966) in German traffic law contexts.
\subsubsection{Narrative-of-Thought (NoT) Prompting}
\label{NoT}
\noindent \textls[-10]{Temporal reasoning remains a significant challenge for LLMs, particularly in inferring global temporal relationships from unordered events. To evaluate this capability,~\cite{zhang2024narrativeofthoughtimprovingtemporalreasoning} introduced Temporal Graph Generation (TGG), a benchmark designed to assess LLMs' proficiency in constructing directed acyclic graphs (DAGs) representing event timelines. Experimental results revealed that smaller LLMs (<10B) lagged behind GPT-3.5/4 by approximately 50\%, with even GPT-4 facing difficulties due to alignment constraints. To overcome these limitations, the authors proposed Narrative-of-Thought (NOT), a prompting strategy that enhances temporal reasoning without requiring additional model training. NOT comprises three core components: (1) Structural Representation, where events are encapsulated in a Python class and processed through code completion; (2) NOT Prompting template, which generates temporally grounded narratives to guide the construction of temporal graphs; and (3) Narrative-Aware Demonstrations, utilizing GPT-4-generated few-shot examples optimized for both conciseness and accuracy. Results demonstrated that NOT significantly improves the performance of small LLMs, with LLaMA3-8B achieving an F1 score of 42.2, closely matching GPT-3.5’s 45.7, while exhibiting superior structural coherence.}
\subsubsection{Buffer of Thoughts (BoT) Prompting}
\label{BoT}
\noindent 
Existing prompting methods often struggle to balance universality, efficiency, and robustness in complex reasoning. To address this,~\cite{yang2024buffer} introduced Buffer of Thoughts (BoT), a framework that enhances LLMs through reusable high-level reasoning patterns.
BoT overcomes the limitations of single-query methods (e.g., manual exemplar reliance) and multi-query approaches (e.g., computational inefficiency) by introducing a meta-buffer that distills "thought-templates" from diverse tasks and a dynamic buffer-manager that continuously refines them as new problems are solved. BoT retrieves task-specific thought-templates (e.g., structured problem-solving approaches) and adaptively instantiates them, mimicking human analogical reasoning to eliminate manual prompt design and recursive exploration. Experiments across 10 benchmarks demonstrate its state-of-the-art performance, achieving gains of 11\% on Game of 24, 20\% on Geometric Shapes, and 51\% on Checkmate-in-One, while using just 12\% of the computational cost of multi-query methods like Tree-of-Thoughts. Notably, BoT enhances smaller models, with Llama3-8B + BoT surpassing Llama3-70B in accuracy, showing its potential to democratize efficient reasoning at scale.
\vspace{-4mm}
\subsubsection{Contrastive Denoising with Noisy Chain-of-Thought (CD-CoT) Prompting}
\label{CD-CoT}
\noindent Contrastive Denoising with Noisy Chain-of-Thought (CD-CoT)~\citep{NEURIPS2024_dfaa29ed} addresses the challenge of "noisy rationales" in chain-of-thought prompting, where irrelevant or incorrect intermediate reasoning steps degrade LLM performance. The NoRa (Noisy Rationales) dataset highlights this issue, showing that LLMs often perform worse with flawed rationales than with no examples at all, as they tend to mimic incorrect reasoning. Existing methods like self-correction and self-consistency offer limited solutions, as self-correction fails without external feedback, and self-consistency selects frequent answers without resolving reasoning flaws. CD-CoT mitigates this by contrasting noisy rationales with clean ones, rephrasing flawed examples, selecting optimal reasoning paths, and voting on the most consistent answer. Experiments show that CD-CoT improves accuracy by 17.8\% on average, significantly outperforming baselines and enhancing LLMs’ robustness in reasoning-intensive tasks.
\subsubsection{Reverse Chain-of-Thought (R-CoT) Prompting}
\label{R-CoT}
\noindent ~\cite{deng2024rcotreversechainofthoughtproblem} introduced the Reverse Chain-of-Thought (R-CoT) pipeline, a novel approach to enhancing geometric reasoning in LMMs by addressing dataset limitations such as low quality, diversity, and fidelity. R-CoT operates in two stages: GeoChain, which generates high-fidelity geometric images with detailed step-by-step descriptions of geometric relationships (e.g., midlines, radii), and Reverse A\&Q, which derives questions from reasoning chains using LLMs, ensuring accurate multi-step problem generation. By prioritizing answer-aware question synthesis, R-CoT mitigates visual hallucinations and reasoning errors in LMMs. The resulting GeoMM dataset includes 20 geometric shapes categorized by complexity, incorporating relational questions often missing in existing datasets like MAVIS and GeomVerse. GeoMM combines high-fidelity images with diverse Q\&A pairs, enriched by geometric theorems and line operations. Experimental results demonstrate that R-CoT-trained models achieve state-of-the-art performance, with the 8B-parameter model surpassing GPT-4o by 12.5\% on MathVista and 14.5\% on GeoQA, while smaller models (2B, 7B) also set new benchmarks.  
\subsubsection{Chain of Draft (CoD) Prompting}
\label{CoD}
\noindent Chain of Draft (CoD)~\citep{xu2025chaindraftthinkingfaster}, a novel prompting strategy designed to enhance efficiency in complex reasoning tasks. Unlike traditional CoT prompting, which emphasizes detailed step-by-step reasoning, CoD generates concise, information-dense outputs at each step, mirroring human problem-solving strategies where only essential insights are noted. While CoT improves reasoning accuracy, it often leads to verbose outputs and increased computational costs. CoD mitigates this by constraining word usage in each reasoning step, reducing latency and token consumption without sacrificing accuracy. This efficiency-oriented approach is particularly valuable for real-world applications where computational resources and response time are critical. Experimental results across arithmetic, commonsense, and symbolic reasoning benchmarks show that CoD matches or even outperforms CoT in accuracy while significantly lowering token usage and latency. In some cases, CoD achieved comparable accuracy with an 80\% reduction in output tokens as well as an average latency reduction of 76.2\%, demonstrating its potential as a lightweight yet effective alternative to traditional prompting strategies.

\subsection{Reduce Hallucination}
\label{rh}
\subsubsection{Retrieval Augmented Generation (RAG)}
\label{RAG}
LLMs have revolutionized text generation, yet their reliance on limited, static training data hinders accurate responses, especially in tasks demanding external knowledge. Traditional prompting falls short, requiring expensive retraining. Retrieval Augmented Generation (RAG)~\citep{lewis2020retrieval} emerges as a novel solution, seamlessly weaving information retrieval into the prompting process. RAG analyzes user input, crafts a targeted query, and scours a pre-built knowledge base for relevant resources. Retrieved snippets are incorporated into the original prompt, enriching it with contextual background. The augmented prompt empowers the LLM to generate creative, factually accurate responses. RAG's agility overcomes static limitations, making it a game-changer for tasks requiring up-to-date knowledge. RAG outperformed seq2seq models and task-specific architectures on ODQA benchmarks, achieving exact match scores, reaching up to 56.8\% on TriviaQA and 44.5\% on Natural Questions.

\subsubsection{ReAct Prompting}
\label{react}
Unlike previous studies that treated reasoning and action separately, ReAct~\citep{yao2022react} enables LLMs to generate reasoning traces and task-specific actions concurrently. This interleaved process enhances synergy between reasoning and action, facilitating the model in inducing, tracking, and updating action plans while handling exceptions. ReAct is applied to diverse language and decision-making tasks, showcasing its effectiveness over state-of-the-art baselines. Notably, in question answering (HotpotQA) and fact verification (Fever), ReAct addresses hallucination and error propagation issues by interacting with a simple Wikipedia API, producing more interpretable task-solving trajectories. Additionally, in interactive decision-making benchmarks like ALFWorld and WebShop, ReAct surpasses both imitation and reinforcement learning approaches, achieving notable success rates of 34\% and 10\%, respectively, with minimal in-context examples.
\subsubsection{Chain-of-Verification (CoVe) Prompting}
\label{cove}
To address hallucinations in LLMs,~\cite{dhuliawala2023chain} proposed Chain-of-Verification (CoVe), which involves a systematic four-step process including the model generate baseline responses, plan verification questions to check its work, answer the questions independently, and produce a revised response incorporating the verification. By verifying its work through this deliberate multi-step approach, the LLM enhances logical reasoning abilities and reduces errors even with contradictory information. CoVe emulates human verification to bolster the coherence and precision of LLM output. Experiments on list questions, QA, and long-form generation demonstrate that CoVe decreases hallucinations while maintaining facts~\citep{sahoo2024comprehensive}. Focused verification questions help models identify and correct their inaccuracies.

\subsubsection{Chain-of-Note (CoN) Prompting}
\label{con}
Retrieval-augmented language models (RALMs) enhance large language models by incorporating external knowledge to reduce factual hallucination. However, the reliability of retrieved information is not guaranteed, leading to potentially misguided responses. Standard RALMs struggle to assess their knowledge adequacy and often fail to respond with "unknown" when lacking information. To address these challenges,~\cite{yu2023chainofnote} introduced a novel approach to improve RALMs robustness by handling noisy, irrelevant documents and accurately addressing unknown scenarios. CoN systematically evaluates document relevance, emphasizing critical and reliable information to filter out irrelevant content, resulting in more precise and contextually relevant responses. Testing across diverse open-domain question-answering datasets demonstrated notable improvements, including a +7.9 average boost in exact match scores for noisy retrieved documents and a +10.5 enhancement in rejection rates for questions beyond pre-training knowledge.

\subsubsection{Chain-of-Knowledge (CoK) Prompting}
\label{cok}
\textls[-10]{Traditional prompting techniques for LLMs have proven powerful in tackling basic tasks. However, their efficacy diminishes due to complex reasoning challenges, often resulting in unreliable outputs plagued by factual hallucinations and opaque thought processes. This limitation arises from their reliance on fixed knowledge sources, ineffective structured query generation, and lack of progressive correction that fails to guide the LLM adequately. Motivated by human problem-solving, CoK~\citep{li2023chainofknowledge} systematically breaks down intricate tasks into well-coordinated steps. The process initiates with a comprehensive reasoning preparation stage, where the context is established, and the problem is framed. Subsequently, it engages in a dynamic knowledge adaptation phase, meticulously gathering evidence from various sources, such as its internal knowledge base, external databases, and the given prompt.}

\subsection{User Interface}
\label{ui}
\subsubsection{Active Prompting}
\label{ap}
\textls[-10]{\cite{diao2023active} introduced Active-Prompting as a solution to the challenge of adapting LLMs to diverse reasoning tasks. They address the issue by proposing Active-Prompt to enhance LLMs' performance on complex question-and-answer tasks through task-specific example prompts with chain-of-thought (CoT) reasoning. Unlike existing CoT methods that rely on fixed sets of human-annotated exemplars, Active-Prompt introduces a mechanism for determining the most impactful questions for annotation. Drawing inspiration from uncertainty-based active learning, the method utilizes various metrics to characterize uncertainty and selects the most uncertain questions for annotation. Active-Prompting exhibits superior performance, outperforming self-consistency by an average of 7.0\% and 1.8\% across eight complex reasoning tasks in text-davinci-002 and code-davinci-002, respectively, showcasing state-of-the-art results. }
\subsection{Fine-Tuning and Optimization}
\label{fto}
\subsubsection{Automatic Prompt Engineer (APE)}
\label{ape}
While crafting effective prompts for LLMs has traditionally been a laborious task for expert annotators, ~\cite{zhou2022large} introduced Automatic Prompt Engineer (APE) as an innovative approach to automatic instruction generation and selection for LLMs. APE sheds the limitations of static, hand-designed prompts by dynamically generating and selecting the most impactful prompts for specific tasks. This ingenious method analyzes user input, crafts candidate instructions, and then leverages reinforcement learning to choose the optimal prompt, adapting it on the fly to different contexts. Extensive tests on the diverse BIG-Bench suite and the CoT reasoning task revealed APE's prowess, exceeding human-authored prompts in most cases (19 out of 24 tasks) and significantly boosting LLMs reasoning abilities. This breakthrough in automatic prompt engineering paves the way for LLMs to tackle a wider range of tasks with greater efficiency and adaptability, unlocking their full potential across diverse applications.

\subsection{Knowledge-Based Reasoning and Generation}\label{kbrg}
\subsubsection{Automatic Reasoning and Tool-use (ART)}
\label{ART}
The limited reasoning abilities and lack of external tool utilization hinder the potential of LLMs in complex tasks.~\cite{paranjape2023art} introduced Automatic Reasoning and Tool-use (ART) to tackle this critical barrier that empowers LLMs to reason through multi-step processes and seamlessly integrate external expertise. ART bridges the reasoning gap, enabling LLMs to tackle complex problems and expand beyond simple text generation. By integrating external tools for specialized knowledge and computations, ART unlocks unprecedented versatility and informs LLM outputs with real-world relevance. This allows LLMs to contribute to diverse fields like scientific research, data analysis, and even decision-making support. Moving beyond traditional prompting techniques, ART automates reasoning steps through structured programs, eliminating the need for laborious hand-crafting. Its dynamic tool integration ensures smooth collaboration, pausing generation to incorporate external tool outputs and seamlessly resuming the flow. Empirical evidence on challenging benchmarks (BigBench and MMLU) demonstrates ART's effectiveness, surpassing traditional prompting and even matching hand-crafted demonstrations in some cases.
\subsection{Improving Consistency and Coherence}
\label{icc}
\subsubsection{Contrastive Chain-of-Thought (CCoT) Prompting}
\label{ccot}
Traditional CoT prompting for LLMs often misses a crucial element: learning from mistakes. That is where Contrastive Chain-of-Thought Prompting (CCoT)~\citep{chia2023contrastive} dives in, providing both valid and invalid reasoning demonstrations alongside original prompts. Imagine exploring a map with the right path and the wrong turns to avoid – that is the advantage of contrastive CoT! This dual-perspective approach, tested on reasoning benchmarks like SQuAD and COPA, pushes LLMs to step-by-step reasoning, leading to 4-16\% improvements in strategic and mathematical reasoning evaluations compared to traditional CoT, further improved by approximately 5\% when integrated with self-consistency techniques. However, questions remain about this technique, such as the automated generation of contrasting demonstrations for diverse problems and its applicability to other NLP tasks beyond reasoning.
\subsection{Managing Emotions and Tone}
\label{met}
\subsubsection{Emotion Prompting} 
\label{EP}
While LLMs demonstrate impressive capabilities on various tasks, their ability to comprehend psychological and emotional cues remains uncertain. The study by~\cite{li2023large} addressed the uncertainty surrounding LLMs' ability to comprehend emotional cues by introducing EmotionPrompt. Drawing inspiration from psychological research on language's impact on human performance, they append 11 emotional stimulus sentences to prompts to enhance LLM emotional intelligence. Experimental results demonstrate seamless integration of these stimuli, significantly improving LLM performance across various tasks. EmotionPrompt demonstrates an 8.00\% relative improvement in instruction induction and an impressive 115\% boost in BIG-Bench tasks, underscoring its efficacy in augmenting LLM capabilities in processing affective signals. An evaluation involving 106 participants reveals an average improvement of 10.9\% in performance, truthfulness, and responsibility metrics for generative tasks when employing EmotionPrompt compared to standard prompts.
\subsection{Code Generation and Execution}
\label{cge}
\subsubsection{Scratchpad Prompting}
\label{scratch}
\textls[-10]{Despite the prowess of Transformer-based language models in generating code for basic programming tasks, they encounter challenges in complex, multi-step algorithmic calculations requiring precise reasoning. Addressing this,~\cite{nye2021show} introduce a novel approach, centered on task design rather than model modification, introduce a `scratchpad' concept. The proposal enables the model to generate an arbitrary sequence of intermediate tokens before providing the final answer. Scratchpad Prompting technique outperforms (Mostly Basic Python Programming) MBPP-aug with a 46.8\% success rate. Combining CodeNet and single-line datasets yields the highest performance, achieving 26.6\% correct final outputs and 24.6\% perfect traces. Scratchpad prompting technique faces limitations, including a fixed context window size of 512 tokens and a dependency on supervised learning for scratchpad utilization.}

\subsubsection{Program of Thoughts (PoT) Prompting}
\label{pot}
\textls[-10]{Language models are suboptimal for solving mathematical expressions due to their proneness to arithmetic errors, incapability in handling complex equations, and inefficiency in expressing extensive iterations. To enhance numerical reasoning in language models, ~\cite{chen2022program} presents Program-of-Thoughts (PoT) prompting, advocating the use of external language interpreters for computation steps. PoT enables models like Codex to express reasoning through executable Python programs, resulting in an average performance improvement of approximately 12\% compared to CoT prompting on datasets involving mathematical word problems and financial questions.}

\subsubsection{Structured Chain-of-Thought (SCoT) Prompting}
\label{scot}
LLMs have exhibited impressive proficiency in code generation. The widely used CoT prompting involves producing intermediate natural language reasoning steps before generating code. Despite its efficacy in natural language generation, CoT prompting demonstrates lower accuracy when applied to code generation tasks.~\cite{li2023structured} introduce Structured Chain-of-Thought (SCoTs) as an innovative prompting technique tailored specifically for code generation. By incorporating program structures (sequence, branch, and loop structures) into reasoning steps, SCoT prompting enhances LLMs' performance in generating structured source code. This approach explicitly guides LLMs to consider requirements from the source code perspective, improving their overall effectiveness in code generation compared to CoT prompting. The authors validated the effectiveness of SCoT on ChatGPT and Codex across three benchmarks (HumanEval, MBPP, and MBCPP) and demonstrated a superior performance over the CoT prompting by up to 13.79\%.

\subsubsection{Chain-of-Code (CoC) Prompting}
\label{ccp}
While CoT prompting has proven very effective for enhancing Language models (LMs) semantic reasoning skills, it struggles to handle questions requiring numeric or symbolic reasoning.~\cite{li2023chain} introduce Chain-of-Code (CoC) as an extension to improve LM reasoning by leveraging codewriting for both logic and semantic tasks. CoC encourages LMs to format semantic sub-tasks as flexible pseudocode, allowing an interpreter to catch undefined behaviors and simulate them with an "LMulator." Experiments demonstrate CoC's superiority over Chain of Thought and other baselines, achieving an 84\% accuracy on BIG-Bench Hard, a 12\% gain. CoC proves effective with both large and small models, expanding LMs' ability to correctly answer reasoning questions by incorporating a "think in code" approach.

\begin{table*}
\centering
\caption{Summary of prevalent prompting techniques of LLMs based on the following factors: application, prompt acquisition, prompt turn, language model, dataset, and metrics.}
\label{table}
\resizebox{0.92\textwidth}{!}{
\begin{tabular}{ccccccc}
\toprule
\multirow{2}{*}{\textbf{Application}} & \multirow{2}{*}{\begin{tabular}[c]{@{}c@{}}\textbf{Prompting}\\ \textbf{Technique}\end{tabular}} & \multicolumn{5}{c}{\textbf{Comparison Scope}} \\
\cmidrule{3-7}
& & \textbf{Prompt Acquisition} & \textbf{Prompt Turn} & \textbf{Language Model(s)}& \textbf{Dataset} &\textbf{Metric(s)} \\
\midrule
\begin{tabular}[c]{@{}c@{}}New Tasks Without \\Training Data\end{tabular} & Zero-shot & Manual & Single & GPT-2 &Arithmetic,Symbolic &Accuracy, ROUGE Score\\
& Few-shot & Manual & Single & GPT-3& NaturalQS, WebQS, TriviaQA& Accuracy\\
\midrule
 & CoT & Manual & Multi & PaLM 540B&GSM8K&Accuracy \\
 & LogiCoT & Manual & Multi & Vicuna-33b, GPT-4&GSM8K, AQuA, SocialQA&Accuracy \\
 & CoS & Manual & Multi & \texttt{gpt-3.5-turbo}, GPT-4 & SPARTUN &Accuracy, Precision, Recall \\
& Auto-CoT & LM Generated & Multi & GPT-3& Arithmetic, Symbolic& Accuracy\\
& Self-Consistency & Manual & Single & PaLM 540B&Arithmetic, Commonsense &Accuracy \\
& ToT & Retrieval Based & Multi & GPT-4& Game of 24, Creative Writing & Success Rate  \\
& GoT & Retrieval Based & Multi & T5-large& GSM8K, ScienceQA& ROUGE Score \\
& S2A & Manual & Single & Llama 2-70B& QA,GSM8K& Accuracy \\
& ThoT & Hybrid & Multi&\texttt{gpt-3.5-turbo}, Llama 2-70b-chat & PopQA, EntityQ, MTCR& Exact Match (EM) Score\\
& Chain of Table & Manual& Multi&GPT 3.5, LLaMA 2 &TabFact, WikiTQ & BLEU, ROUGE Score \\
\begin{tabular}[c]{@{}c@{}}Reasoning and Logic\end{tabular} & Self-Refine & Manual & Multi& GPT‑3.5,GPT‑4 &7 diverse tasks(e.g., Dialogue Response,Math Reasoning) &Task‑specific (Accuracy, Human Preference)  \\
& Code Prompting & LM Generated & Multi& GPT 3.5, Mixtral &CondQA,ShaRC,BGQA & F1 \\
& ECHO & Hybrid & Multi& \texttt{gpt-3.5-Turbo-0301} &Arithmetic,Commonsense,Symbolic& Accuracy \\
& Logic-of-thought & LM Generated & Multi& GPT 3.5-turbo, GPT-4 &ReClor, LogiQA, RuleTaker, ProofWriter, FOLIO & Accuracy \\
& IAP & Manual& Multi& LLaMA-3-8B-Instruct, Qwen-14B-Chat &Math,Logic,Commonsense & Accuracy \\
& EEDP & Manual & Single &GPT-4-turbo &Merged 1000, ZINC test 2500 & Accuracy \\
& LoT & LM Generated & Multi&GPT-4o &Japanese Civil Law,Normative sentence & Precision,Recall,F2 \\
& NoT & LM Generated & Single & GPT-3.5,GPT-4, Mistral-7B,LLaMA3-8B &ProScript,Schema-11,WikiHow Script & F1,GED \\
& BoT & LM Generated  & Multi &Llama3‑8B, Llama3‑70B  & 10 reasoning‑intensive tasks (e.g., Game of 24, Geometric Shapes)& Accuracy \\
& CD-CoT & Manual & Single& \texttt{gpt-3.5-turbo-0613}, Gemini-Pro(and others)  & Multiple tasks (e.g. BIG‑Bench subsets, commonsense QA, etc.)&Accuracy, Solve Rate, Human Preference \\
& R-CoT & Manual &Single & GPT4o, R-CoT-8B(and others) & GeoMM,MathVista,GeoQA& Accuracy\\
& CoD & Hybrid & Single&GPT-4o,Claude 3.5
Sonnet  & Arithmetic, Commonsense, Symbolic&Accuracy \\

\midrule
 & CoVe & Retrieval Based & Multi & Llama 65B &Wikidata, QUEST, MultiSpanQA& Precision, F1 \\
& ReAct & Retrieval Based & Multi & PaLM-540B, GPT-3& HotpotQA, FEVER&Exact Match (EM), Accuracy   \\
\begin{tabular}[c]{@{}c@{}}Reduce Hallucination\end{tabular} & RAG & Retrieval Based & Single & RAG-Token, RAG-Seq. &MSMARCO, SearchQA &ROUGE, BLEU score \\
& CoN & LM Generated & Multi & Llama 2, DPR& NQ, TriviaQA, WebQ &Exact Match (EM), F1 Score \\
& CoK & LM Generated & Multi & \texttt{gpt-3.5-turbo-0613} & \begin{tabular}[c]{@{}c@{}}HotpotQA, FEVER, MedMCQA, \\MMLU Physics and Biology\end{tabular} &Exact Match (EM), Accuracy \\
\midrule
\begin{tabular}[c]{@{}c@{}}User Interaction\end{tabular} & Active-Prompt & Manual & Single & \texttt{code-davinci-002}, \texttt{text-davinci-003}& Arithmetic, Commonsense, Symbolic & \begin{tabular}[c]{@{}c@{}}Disagreement, Entropy\\ Variance, Self-confidence Score\end{tabular}\\
\midrule
\begin{tabular}[c]{@{}c@{}}Fine-Tuning and\\ Optimization\end{tabular} & APE & LM Generated & Single & \texttt{text-curie-001}, \texttt{text-davanci-002} & BBII, TruthfulQA & \begin{tabular}[c]{@{}c@{}}Execution accuracy, Log probability,\\ Efficient score estimation\end{tabular}  \\
\midrule
\begin{tabular}[c]{@{}c@{}}Knowledge-Based \\Reasoning and Generation\end{tabular} & ART & Hybrid & Multi & GPT-3 (175B)& BigBench, MMLU & Accuracy \\
\midrule
\begin{tabular}[c]{@{}c@{}}Improving Consistency\\ and Coherence\end{tabular} & CCoT & LM Generated & Multi & \texttt{gpt-3.5-turbo-0301}& Arithmetic, Factual QA& Accuracy\\
\midrule
\begin{tabular}[c]{@{}c@{}}Managing Emotions \\and Tone\end{tabular} & Emotion Prompting & Manual & Single & GPT-4 &BIG-Bench, Instruction Induction& Accuracy \\
\midrule
& SCoT & Hybrid & Multi & ChatGPT, Codex& HumanEval, MBPP, MBCPP&pass@$$k$$ \\
\begin{tabular}[c]{@{}c@{}}Code Generation \\and Execution\end{tabular} & PoT & Manual & Single & \texttt{gpt-3.5-turbo}& GSM8K, SVAMP, FinQA& Exact Match(EM) Score \\
& CoC & Manual & Single & \texttt{text-davinci-003}, \texttt{gpt-3.5-turbo} & BIG-Bench Hard & Accuracy \\
& Scratchpad Prompting & Manual & Single & GPT-3 & MBPP, MBPP-aug& Accuracy \\
\midrule
\begin{tabular}[c]{@{}c@{}}Optimization and\\ Efficiency\end{tabular} & OPRO & Manual & Single & PaLM 2-L-IT, text-bison& GSM8K, BIG-Bench Hard& Accuracy \\
\midrule
\begin{tabular}[c]{@{}c@{}}Understanding \\ User Intent\end{tabular} & RaR& Manual & Single &\texttt{GPT-4-061}3& Knowledge, Symbolic& \begin{tabular}[c]{@{}c@{}}Accuray, Fair Score,\\Language Modeling Score\end{tabular}\\
\midrule
\begin{tabular}[c]{@{}c@{}}Metacognition \\ and Self-Reflection\end{tabular} & Take a Step Back&Manual & Single&PaLM2-L, GPT-4& \begin{tabular}[c]{@{}c@{}}MMLU-Physics, 
MMLU-Chemistry\\ TimeQA, SituatedQA, StrategyQA\end{tabular}
 & Accuracy \\
\bottomrule
\end{tabular}
}
\end{table*}

\subsection{Optimization and Efficiency}
\label{oe}
\subsubsection{Optimization by Prompting (OPRO)}
\label{opro}
In various domains, optimization is a fundamental process often involving iterative techniques. \cite{yang2023large} introduce Optimization by PROmpting (OPRO), a novel approach that leverages LLMs as optimizers. Unlike traditional methods, OPRO utilizes natural language prompts to iteratively generate solutions based on the problem description, enabling quick adaptation to different tasks and customization of the optimization process. The potential of LLMs for optimization is demonstrated through case studies on classic problems like linear regression and the traveling salesman problem. Additionally, it explores the optimization of prompts to maximize accuracy in natural language processing tasks, highlighting the sensitivity of LLMs. The experiments show that optimizing prompts for accuracy on a small training set effectively translates to high performance on the test set. OPRO leads to a significant performance boost, with the most effective prompts optimized by OPRO outperforming human-designed prompts by up to 8\% on the GSM8K dataset and up to 50\% on challenging tasks in Big-Bench.

\subsection{Understanding User Intent}
\label{uui}
\subsubsection{Rephrase and Respond (RaR) Prompting}
\label{RaR}
The study by~\cite{deng2023rephrase} brings attention to an often-neglected dimension in exploring LLMs: the disparity between human thought frames and those of LLMs and introduces Rephrase and Respond (RaR). RaR allows LLMs to rephrase and expand questions in a single prompt, demonstrating improved comprehension and response accuracy. The two-step RaR variant, incorporating rephrasing and response LLMs, achieves substantial performance enhancements across various tasks. The study highlights that in contrast to casually posed human queries, the rephrased questions contribute to enhanced semantic clarity and the resolution of inherent ambiguity. These findings offer valuable insights for understanding and enhancing the efficacy of LLMs across various applications.

\subsection{Metacognition and Self-Reflection}
\label{mas}
\subsubsection{Take a Step Back Prompting}
\label{tsbp}
Addressing the persistent challenge of complex multi-step reasoning, \cite{zheng2023take} introduced the Step-Back prompting technique, tailored explicitly for advanced language models like PaLM-2L. This innovative approach empowers models to engage in abstraction, extracting high-level concepts and fundamental principles from specific instances. The Step-Back prompting method involves a two-step procedure, integrating Abstraction and Reasoning. Through extensive experiments, applying Step-Back Prompting to PaLM-2L in diverse reasoning-intensive tasks such as STEM, Knowledge QA, and Multi-Hop Reasoning, the results demonstrate a substantial enhancement in reasoning capabilities. Noteworthy performance boosts are observed, with improvements in tasks like MMLU Physics and Chemistry by 7\%, TimeQA by 27\%, and MuSiQue by 7\%.

\section{Conclusion}
In the domain of artificial intelligence, prompt engineering has become a transformative force, unlocking the vast potential of LLMs. This survey paper aims to serve as a foundational resource that systematically categorizes 41 distinct prompt engineering techniques based on their targeted functionalities, inspiring further research and empowering innovators in the evolving landscape of prompt engineering. The analysis spans applications, models, and datasets, shedding light on the strengths and limitations of each approach. Furthermore, we have added a diagram and a table to highlight the important points. Despite the remarkable successes, challenges persist, including biases, factual inaccuracies, and interpretability gaps, necessitating further investigation and mitigation strategies. The future of prompt engineering holds immense potential, with emerging trends like meta-learning and hybrid prompting architectures promising amplified capabilities. However, ethical considerations are paramount, emphasizing responsible development and deployment to ensure positive integration into our lives.

\bibliographystyle{named}
\bibliography{ijcai24}

\end{document}